# Multi-label topic classification for COVID-19 literature with Bioformer

Li Fang[1], Kai Wang[1, 2*]

[1] Raymond G. Perelman Center for Cellular and Molecular Therapeutics,

Children's Hospital of Philadelphia, Philadelphia, PA 19104, USA.

[2] Department of Pathology and Laboratory Medicine, University of Pennsylvania

Perelman School of Medicine, Philadelphia, PA 19104, USA.

[*]Correspondence: wangk@chop.edu

# Abstract

We describe Bioformer team's participation in the multi-label topic classification task for COVID-19 literature (track 5 of BioCreative VII). Topic classification is performed using different BERT models (BioBERT, PubMedBERT, and Bioformer). We formulate the topic classification task as a sentence pair classification problem, where the title is the first sentence, and the abstract is the second sentence. Our results show that Bioformer outperforms BioBERT and PubMedBERT in this task. Compared to the baseline results, our best model increased micro, macro, and instance-based F1 score by 8.8%, 15.5%, 7.4%, respectively. Bioformer achieved the highest micro F1 and macro F1 scores in this challenge. In post-challenge experiments, we found that pretraining of Bioformer on COVID-19 articles further improves the performance.

# Introduction

Since the initial outbreak of the coronavirus disease 2019 (COVID-19), there has been an explosion of new scientific literature(1). LitCovid is a curated literature resource of COVID-19 studies (2,3). LitCovid is updated daily, and the new articles are curated into eight topic categories including mechanism, transmission, diagnosis, treatment, prevention, case report, forecasting and general. An automated topic classification pipeline can greatly help the curation process. Track 5 of BioCreative VII calls for a community effort to develop novel methods for this topic classification problem (4). In this task, each COVID-19-related article can be classified into one or more categories.

A transformer model is a deep learning model with self-attention mechanisms (5). The original transformer model was a sequence-to-sequence model and greatly improved the performance of machine translation (5). Bidirectional Encoder Representations from Transformers (BERT) is a transformer-based model and is pretrained on two tasks: masked language modeling and next sentence prediction (6). Thanks to this pretraining process, BERT learns contextual embeddings of words. BERT and its variants (e.g. RoBERTa (7)) have brought significant performance gains on a variety of language tasks. BERT has been adapted to the biomedical domain (8-10). Recently, we pretrained a compact biomedical BERT model named Bioformer. In this study, we focus on solving the multi-label topic classification problem using Bioformer and comparing its performance with other two biomedical BERT models (BioBERT (8) and PubMedBERT (10)). All the three BERT models provide significant performance increases compared to the baseline methods while Bioformer performs the best. Bioformer achieved the highest micro F1 and macro F1 scores in this challenge (4).

# Materials and Methods

## Training, development and testing data set

The training and development set of the task contain 24,960 and 6,239 articles, respectively. The testing set contains 2,500 articles. Each article has the information of journal name, article title, abstract, keywords (optional), publication type, authors, and DOI. Unlike the categories from the LitCovid website, the specific task in this challenge does not include the "General" category and only has seven categories: Mechanism, Transmission, Diagnosis, Treatment, Prevention, Case Report, and Epidemic Forecasting.

## Models used in this study

We used BioBERT(8), PubMedBERT(10) and Bioformer (https://github.com/WGLab/bioformer/) to perform the multi-label topic classification. For BioBERT, we used BioBERT$_{Base-v1.1}$, which is the version described in the publication. PubMedBERT has two versions: one version was pre-trained on PubMed abstracts (denoted by PubMedBERT$_{Ab}$ in this study), and the other version was pre-trained on PubMed abstracts plus PMC full texts (denoted by PubMedBERT$_{AbFull}$). We used Bioformer$_{8L}$ which is a compact Biomedical BERT model with 8 hidden layers. Bioformer$_{8L}$ was pretrained on PubMed abstracts and one million PMC full-text articles for 2 million steps.

## Topic classification

We formulate the topic classification task as a sentence pair classification problem, where the title is the first sentence and the abstract is the second sentence. The input is represented as "[CLS] title [SEP] abstract [SEP]". The representation of the [CLS] token in the last layer was used to classify the relations. We utilized the sentence classifier in transformers python library to fine-tune the models. We treated each topic independently and fine-tuned seven different models (one per topic). We fine-tuned each BERT model on the training dataset for 3 epochs. The maximum input sequence length was fixed to 512. A batch size of 16 was selected, and a learning rate of 3e-5 was selected.

## Further pretraining Bioformer on COVID-19 articles

We downloaded the abstracts of 164,179 COVID-19 articles from the LitCovid website (accessed on Aug 23, 2021), and the total size of the abstracts was 164MB. The pretraining was performed on Google Colab with TPU (v2-8) acceleration. The max input length is fixed to 512 and the batch size was set to 256. The learning rate is set to 2e-5. We pretrained Bioformer on this dataset for 100 epochs where each epoch has different random masking positions. The number of optimization steps is about 80k. The pretraining was finished in 8 hours. We denote this model as Bioformer$_{LitCovid}$.

# Results

## Performance on the development set

The performance on the development set is shown in Table 1. The performance was evaluated using the script provided by the challenge organizer. As this is a multi-label classification task, four different average F1 scores are presented. Bioformer$_{8L}$ achieves best performance on three metrics: instance-based F1, weighted average F1, and micro F1. Both PubMedBERT$_{Ab}$ and PubMedBERT$_{AbFull}$ performer better than BioBERT$_{Base-v1.1}$.

Table 1. Development Set Performance

| Model | Micro F1 | Macro F1 | Instance-based F1 | Weighted average F1 |
|---|---|---|---|---|
| Bioformer$_{8L}$ | 91.05 (1) | 86.60 (2) | 91.69 (1) | 91.06 (1) |
| PubMedBERT$_{AbFull}$ | 90.89 (2) | 86.44 (3) | 91.68 (2) | 90.90 (2) |
| PubMedBERT$_{Ab}$ | 90.80 (3) | 86.62 (1) | 91.49 (3) | 90.82 (3) |
| BioBERT$_{Base-v1.1}$ | 90.77 (4) | 86.14 (4) | 91.47 (4) | 90.77 (4) |

Note: F1 scores are scaled by 100x. The number in the parentheses indicates the ranking of the model.

## Performance on the test set

We submitted the prediction results of five fine-tuned models (described in Table 2). These include three models (Bioformer$_{8L}$, PubMedBERT$_{Ab}$, and BioBERT$_{Base-v1.1}$) fine-tuned on the training set and one model (Bioformer$_{8L}$) fine-tuned on the combination of training and development set. As of Sep 12$^{th}$, 2021, the LitCovid website provided more than 164,000 articles with labeled topics. To test if we can get a performance gain from this information, we fine-tuned Bioformer$_{8L}$ on all articles from the LitCovid website based on their labels (denoted as Bioformer$_{8L}$-web).

Table 2. Description of Submitted Models

| Fine-tuned model name | Pretrained Model | Fine-tuning data |
|---|---|---|
| Bioformer$_{8L}$-train | Bioformer$_{8L}$ | training set |
| PubMedBERT$_{Ab}$-train | PubMedBERT$_{Ab}$ | training set |
| BioBERT$_{Base-v1.1}$-train | BioBERT$_{Base-v1.1}$ | training set |
| Bioformer$_{8L}$-train-dev | Bioformer$_{8L}$ | training + dev set |
| Bioformer$_{8L}$-web | Bioformer$_{8L}$ | LitCovid website |

The results on the testing set returned by the challenge organizer are shown in Table 3. We also showed the baseline performance (11) and team statistics. We first compared the three models that were fined-tuned on the same dataset (the official training set). Similar to the development set results, Bioformer$_{8L}$ outperforms PubMedBERT$_{Ab}$ and BioBERT$_{Base-v1.1}$ in terms of micro F1 and instance-based F1. PubMedBERT$_{Ab}$ achieved better macro F1 than the other two models. Fine-tuning on the combination of training and development set improved the micro F1 score, which is often the preferred metric for multi-class classification when there is class imbalance. Fine-tunning on labeled articles from the LitCovid website (Bioformer$_{8L}$-web) failed to improve the performance. After the challenge, we learned that not all articles in the LitCovid website are manually curated. It includes a substantial portion of articles that were classified by text-mining tools. All our submissions provide significant performance gain compared with the baseline method. Our best model (Bioformer$_{8L}$-train-dev) increased micro, macro and instance-based F1 by 8.8%, 15.5%, 7.4%, respectively.

Table 3. Test Set Performance

| Model | Micro F1 | Macro F1 | Instance-based F1 |
| --- | --- | --- | --- |
| Bioformer$_{8L}$-train-dev | 91.81 (1) | 88.39 (4) | 93.24 (2) |
| Bioformer$_{8L}$-train | 91.79 (2) | 88.70 (2) | 93.34 (1) |
| BioBERT$_{Base-v1.1}$-train | 91.70 (3) | 88.63 (3) | 93.14 (3) |
| PubMedBERT$_{Ab}$-train | 91.66 (4) | 88.75 (1) | 93.11 (4) |
| Bioformer$_{8L}$-web | 90.35 (5) | 87.43 (5) | 91.69 (5) |
| Baseline (ML-Net)(11) | 84.37 | 76.55 | 86.78 |
| Mean of all teams | 87.78 | 81.91 | 89.31 |
| Q1 of all teams | 85.41 | 76.51 | 86.68 |
| Median of all teams | 89.25 | 85.27 | 91.32 |
| Q3 of all teams | 90.83 | 86.70 | 92.54 |

Note: F1 scores are scaled by 100x. The number in the parentheses indicates the ranking in our five submissions (rather than the ranking among all teams).

## Pretraining of Bioformer on COVID-19 articles improves the performance

Bioformer was pretrained on all PubMed abstracts and a fraction of PubMed Central full-text articles. To test if a more specific pretraining corpus could improve the performance, we pretrained Bioformer on abstracts of COVID-19 articles for 100 epochs (see Method sections for details). The new model was denoted as Bioformer$_{Litcovid}$. We then fine-tuned Bioformer$_{Litcovid}$ on the training set of the challenge. We repeated the fine-tuning process for five times using different seeds. The pretraining process was finished before the challenge, but the repeat of fine-tuning experiments was performed after the challenge. The results are shown in Table 4. Bioformer$_{Litcovid}$ has better performance in all three metrics, indicating that pretraining on a more specific corpus is beneficial to the downstream task.

Table 4. Performance of Bioformer$_{8L}$ and Bioformer$_{Litcovid}$ on the development set

| seed | micro F1 | | macro F1 | | instance-based F1 | |
|---|---|---|---|---|---|---|
| | Bioformer$_{8L}$ | Bioformer$_{Litcovid}$ | Bioformer$_{8L}$ | Bioformer$_{Litcovid}$ | Bioformer$_{8L}$ | Bioformer$_{Litcovid}$ |
| 20211125 | 90.92 | 90.93 | 86.46 | 86.68 | 92.77 | 92.85 |
| 20211126 | 90.96 | 90.94 | 86.77 | 86.74 | 92.81 | 92.76 |
| 20211127 | 90.89 | 90.96 | 86.61 | 86.73 | 92.77 | 92.87 |
| 20211128 | 90.72 | 90.79 | 86.12 | 86.27 | 92.72 | 92.70 |
| 20211129 | 90.89 | 90.85 | 86.40 | 86.31 | 92.72 | 92.71 |
| Average | 90.88 | 90.89 (+0.01) | 86.47 | 86.55 (+0.08) | 92.76 | 92.78 (+0.02) |

## Performance of single sentence classification

Our original method formulated the topic classification problem as a sentence pair classification task where the title was the first sentence and the abstract was the second sentence. The input was represented as "[CLS] title [SEP] abstract [SEP]". The two sentences were separated by a special token ("[SEP]") and had different segment embeddings. We are curious about the performance of a single sentence classification method that concatenates the title and the abstract as a single sentence input. To answer this question, we fine-tuned Bioformer$_{Litcovid}$ on the training set using this single sentence classification method. The performance on development set is shown in Table 5. Compared with the sentence pair method, the single sentence method has lower micro F1 and macro F1 scores, but has a slightly higher instance-based F1 score.

Table 5. Performance comparison of sentence pair classification and single sentence classification

| seed | micro F1 | | macro F1 | | instance-based F1 | |
|---|---|---|---|---|---|---|
| | sentence pair | single sentence | sentence pair | single sentence | sentence pair | single sentence |
| 20211125 | 90.93 | 90.90 | 86.68 | 86.23 | 92.85 | 92.82 |
| 20211126 | 90.94 | 90.94 | 86.74 | 86.66 | 92.76 | 92.83 |
| 20211127 | 90.96 | 90.82 | 86.73 | 86.61 | 92.87 | 92.76 |
| 20211128 | 90.79 | 90.90 | 86.27 | 86.40 | 92.70 | 92.86 |
| 20211129 | 90.85 | 90.77 | 86.31 | 86.23 | 92.71 | 92.67 |
| Average | 90.89 | 90.87 (-0.02) | 86.55 | 86.43 (-0.12) | 92.78 | 92.79 (+0.01) |

Note: Both classification methods are based on Bioformer$_{Litcovid}$.

## Discussion

In this paper, we present Bioformer team's approaches for the LitCovid Multi-label Topic Classification Track. Our results show that Bioformer outperforms two other BERT models in this task. Our best model provides significant performance gain compared to the baseline method. In post challenge experiments, we showed that further pretraining of Bioformer on COVID-19 articles improved the performance on all three metrics (micro F1, macro F1 and instance-based

F1), indicating the beneficial effects of a more specific corpus. We also tried the single sentence classification method, which didn't improve the performance.

There are some caveats of the current results that we wish to discuss. First, BioBERT and PubMedBERT was released before the outbreak of the COVID-19 and therefore COVID-19 studies were not included in the pretraining corpus of the two models. Bioformer was pretrained in Feb 2021 and its pretraining corpus contains 95,185 COVID-19 studies (0.3% of the total corpus) published before Feb 2021. This fact may partially explain why Bioformer achieved a better performance with fewer number of parameters. Second, Bioformer was pretrained for 2 million steps which is twice as many as that of BioBERT, and the additional training may also contribute to improved performance. In summary, our study demonstrated that a lightweight model Bioformer can achieve satisfactory performance in topic classification on COVID-19 articles. We hope our study facilitate automated topic classification task for scientific literature beyond COVID-19 articles.

# Acknowledgements

This study is partly supported by the Google TPU Research Cloud (TRC) program and NIH/NLM grant LM012895. We thank all the authors and researchers on COVID-19 in making their scientific discoveries available through publications and preprints, and thank the LitCovid developers in providing a platform for query and exchange of scientific information. We would like to thank the organizers of the BioCreative VII track 5 for organizing the challenge to evaluate different informatics tools for text analysis.

# Data Availability

The pretrained model Bioformer$_{Litcovid}$ is publicly available on HuggingFace: https://huggingface.co/bioformers/bioformer-litcovid .

# Competing interests

The authors declare that they have no competing interests.